\documentclass{article}
\usepackage{spconf,amsmath,graphicx}

\usepackage{enumitem}

\usepackage{url}

\usepackage{hyperref} 

\usepackage{multirow}

\DeclareMathOperator*{\argmax}{argmax}

\title{Forgetful Active Learning with Switch Events: Efficient Sampling for Out-of-Distribution Data}
\name{Ryan Benkert, Mohit Prabhushankar, and Ghassan AlRegib}
\address{
OLIVES at the Center for Signal and Information Processing,\\
School of Electrical and Computer Engineering,\\
Georgia Institute of Technology,\\
Atlanta, GA,  30332-0250, USA \\
\{rbenkert3, mohit.p, alregib\}@gatech.edu}
\begin{document}
%\ninept
%
\onecolumn % make sure you keep this coverpage as one column. In this template, we force the coverpage to be one column with this command and then switch to double column for the remaining of the paper with the \doublecolumn command. 

\begin{description}[labelindent=-1cm,leftmargin=1cm,style=multiline]

\item[\textbf{Citation}]{R. Benkert, M. Prabhushankar, and G. AlRegib, “Forgetful Active Learning With Switch Events: Efficient Sampling for Out-of-Distribution Data,” in IEEE International Conference on Image Processing (ICIP), Bordeaux, France, Oct. 16-19 2022} \\

% \item[\textbf{DOI}]{\url{https://doi.org/10.1109/MSP.2017.2783449}}

\item[\textbf{Review}]
{
Date of acceptance: June 2022
} \\

% \item[\textbf{Data and Codes}]{\url{https://ghassanalregib.com/vip-cup/}}% If you do not have data related to this paper, you can remove the data keyword.
%\item[\textbf{Codes}]{\url{https://github.com/olivesgatech/Patient-Aware-Active-Learning}} \\

\item[\textbf{Bib}] {@ARTICLE\{benkert2022\_ICIP,\\ 
author=\{R. Benkert, M. Prabhushankar, and G. AlRegib\},\\ 
journal=\{IEEE International Conference on Image Processing\},\\ 
title=\{Forgetful Active Learning With Switch Events: Efficient Sampling for Out-of-Distribution Data\}, \\ 
year=\{2022\}\\ 
% volume=\{35\},\\ 
% number=\{2\},\\ 
% pages=\{154-161\},\\ 
% keywords=\{traffic engineering computing;video signal processing;IEEE Video and Image Processing Cup 2017 Student Competition;traffic signs\},\\ 
% doi=\{10.1109/MSP.2017.2783449\},\\ 
% ISSN=\{1053-5888\},\\ 
% month=\{March\},\}
} \\

% Preprint sharing policy can vary depending on the publisher. Before posting a paper to arXiv, please specifically check the transaction/convference you are targeting. In some transactions, papers are usually added to arxiv after acceptance. Pubslishers usually allow the authors to share accepted version of their papers but not the final formatted version that is provided by the pubisher. In case of sharing preprints, publishers usually ask to add DOI and citation to the paper along with a copyright notice.

\item[\textbf{Copyright}]{\textcopyright 2022 IEEE. Personal use of this material is permitted. Permission from IEEE must be obtained for all other uses, in any current or future media, including reprinting/republishing this material for advertising or promotional purposes,
creating new collective works, for resale or redistribution to servers or lists, or reuse of any copyrighted component
of this work in other works. }
\\
\item[\textbf{Contact}]{\href{mailto:rbenkert3@gatech.edu}{rbenkert3@gatech.edu}  OR \href{mailto:alregib@gatech.edu}{alregib@gatech.edu}\\ \url{http://ghassanalregib.info/} \\ }
\end{description}

%Following command sequence was used to start the paper content from the following page and avoid numbering cover page.
\thispagestyle{empty}
\newpage
\clearpage
\setcounter{page}{1}

%Cover page was 1 column. \twocolumn changes the page format back to double column.
\twocolumn
\maketitle
\begin{abstract}
This paper considers deep out-of-distribution active learning. In practice, fully trained neural networks interact randomly with out-of-distribution (OOD) inputs and map aberrant samples randomly within the model representation space. Since data representations are direct manifestations of the training distribution, the data selection process plays a crucial role in outlier robustness. For paradigms such as active learning, this is especially challenging since protocols must not only improve performance on the training distribution most effectively but further render a robust representation space. However, existing strategies directly base the data selection on the data representation of the unlabeled data which is random for OOD samples by definition. For this purpose, we introduce \emph{forgetful active learning with switch events} (\texttt{FALSE}) - a novel active learning protocol for out-of-distribution active learning. Instead of defining sample importance on the data representation directly, we formulate "informativeness" with learning difficulty during training. Specifically, we approximate how often the network ``forgets" unlabeled samples and query the most ``forgotten" samples for annotation. We report up to 4.5\% accuracy improvements in over 270 experiments, including four commonly used protocols, two OOD benchmarks, one in-distribution benchmark, and three different architectures. 
\end{abstract}
\section{Introduction}
A major issue in neural network deployment is their sensitivity to unknown inputs \cite{hendrycks2019robustness, kwon2020backpropagated, lee2020gradients, kwon2020novelty}. Within the context of deep learning, unknown inputs typically refer to samples originating from acquisition setups that significantly differ from the training environment. Due to the probabilistic nature of neural network predictions, samples originating from the training environment are called in-distribution while unknown inputs are called out-of-distribution, or OOD in short. In practice, OOD samples result in unpredictable or even random predictions and represent a major challenge for real-world deployment~\cite{prabhushankar2021contrastive, temel2017cure}. The root cause of the perceived unpredictability lies within the structure of the model representation~\cite{lehman2020structures}. During training, the structure is established by imposing constraints (through e.g. a loss function) on known training samples and the training distribution is mapped to distinguishable targets. During test phase, the model maps known in-distribution samples to their respective targets if they originate from the same training distribution. In contrast, OOD samples are not constrained during training and the target representation is undefined. This results in unpredictable behavior at the test phase.

\begin{figure}[!th]
    \begin{center}
        \includegraphics[scale=0.36]{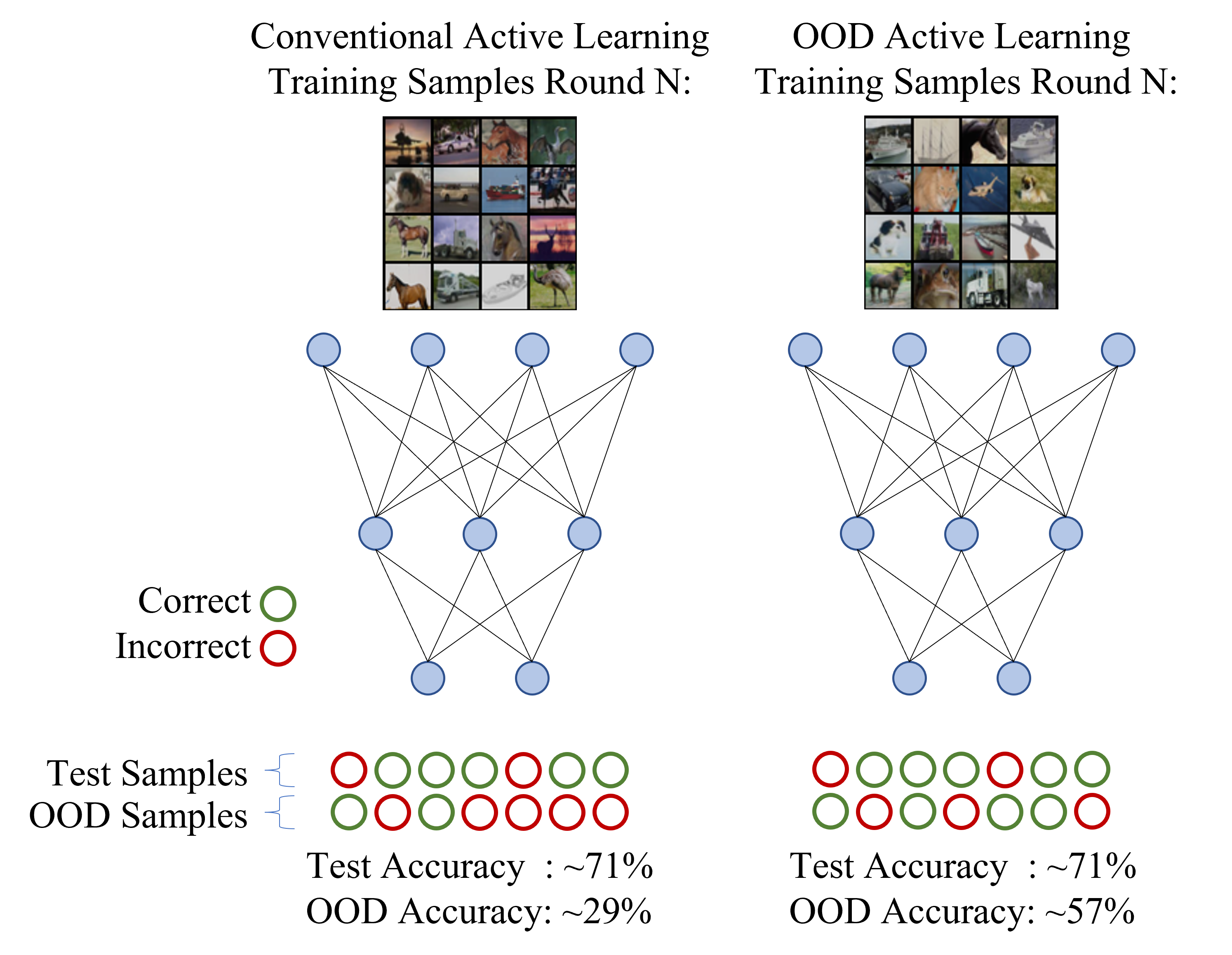}
    \end{center}
    \caption{Toy example of out-of-distribution (OOD) samples within an active learning setting. In both active learning protocols the model exhibits similar test performance. However, the accuracy on the out-of-distribution samples significantly improves when the training set is selected with a different protocol.}
 \label{fig:ood-al-toy}
\end{figure}

A field in which OOD properties are especially important is active learning. Active learning is a machine learning paradigm in which the model iteratively selects the training set from an unlabeled data pool to improve annotation efficiency. Due to its intuitive practicality, active learning is especially popular in applications where annotations are costly and has been successfully deployed in various industrial sectors \cite{ahmadal}. In active learning, out-of-distribution performance is especially relevant because it is not directly apparent from the target performance. For instance, a model may perform significantly better on OOD data if the training set contains samples that impose constraints on outliers (Figure~\ref{fig:ood-al-toy}). For this purpose, active learning protocols must not only maximize performance on in-distribution data but further insure robustness on outlier distributions.

Despite the high relevance for the field, existing active learning approaches rarely consider OOD settings. In fact, most strategies select the next set of training samples based on importance metrics \emph{directly} derived from the model representation. Since model representations for OOD data are undefined by definition, this trait poses a clear inefficiency for OOD test performance. For this purpose, we introduce \texttt{FALSE} - a novel query strategy that not only shows favorable performance on in-distribution data but is further robust to OOD samples. Instead of directly defining sample importance on model representations directly, we define importance with statistics gathered from the model during training. Specifically, we approximate how often unlabeled samples are ``forgotten" by the model and consider the ``most forgotten" samples as the most informative. We benchmark our method against four commonly used active learning protocols on two popular OOD benchmarks, as well as one in-distribution benchmark with three architectures. Overall, we note a significant improvement in terms of test accuracy on in-distribution, as well as out-of-distribution data.

%For dynamic paradigms such as active learning, the unkown out-of-distribution property can have severe implications. For instance, a model may show acceptable performance for in-distribution data but could be significantly improved for OOD samples if several additional aberrant samples are added to the training set. For this purpose, we reason that query strategies must not only generalize well on in-distribution data but also be robust to different outlier distributions. Despite the high relevance, existing strategies do not consider out-of-distribution properties. In fact, most strategies 

%Examples for successful deployment include manufacturing \cite{tong2001active}, robotics \cite{alrobotics}, recommender systems \cite{alrecommender}, medical imaging \cite{hoi2006batch}, and even object detection in autonomous vehicles \cite{haussmann2020scalable}.
\section{Background and Related Work}
\subsection{Active learning}
In active learning \cite{cohn1996active}, the goal is to maximize generalization performance with minimal data annotations. We consider a dataset $D$, an initial training set $D_{train}$, as well as an unlabeled data pool $D_{pool}$ from which we select the next sample batch for the training set. For batch active learning, the algorithm selects a batch of $b$ samples $X^* = \{x^*_1, ..., x^*_b\}$ that are the most informative based on a predefined definition. The selection of $X^*$ is called the query strategy and typically queries the batch that maximizes the acquisition function $a(x_1, ..., x_b | f_w)$, where $f_w$ represents the deep neural network with parameters $w$. In active learning, the definition of importance is encoded within the acquisition function $a$. In this regard, typical approaches involve querying samples that are the most difficult for generalization \cite{wang2014new, houlsby2011bayesian}, maximize data diversity \cite{sener2017active, gissin2019discriminative}, or perform a mixture of generalization difficulty and data diversity \cite{ash2019deep, hsu2015active}. Even though a large variety of strategies exist, they define sample importance on the representation manifold directly. Since representations are notoriously unpredictable for OOD data this characteristic represents a clear inefficiency. Apart from rare exceptions \cite{kothawade2021similar}, out-of-distribution active learning settings are not considered in existing literature.  

%As an example for the different definition types, \cite{wang2014new} query the samples with the highest softmax entropy, \cite{sener2017active} maximize data diversity by constructing the core-set of the unlabeled pool, and \cite{hsu2015active} propose a bandit-style approach that switches between different strategies each round. Even though a large variety of strategies exist, they define sample importance on the representation manifold directly. Since representations are notoriously unpredictable for out-of-distribution data this characteristic represents a clear inefficiency. To the best of our knowledge, out-of-distribution active learning settings are not considered in existing literature.

\subsection{Forgetting Events}
Our approach most closely relates to the study of continual learning or more specifically catastrophic forgetting \cite{toneva2018empirical, ritter2018online, kirkpatrick2017overcoming, benkert2021explaining}. While several papers study forgetting in the context of different target tasks \cite{ritter2018online, kirkpatrick2017overcoming}, other papers focus on forgetting within a single task \cite{toneva2018empirical}.
Within the context of active learning, we define informativeness with learning difficulty. Specifically, we say a sample is informative if it was ``forgotten" frequently during the training process and therefore difficult to learn. Within the context of deep learning, a sample is considered ``forgotten" if it was correctly classified (``learned") at time $t$ and misclassified (``forgotten") at a later time $t' > t$. More formally, we consider recognition tasks where the model calculates a prediction $\Tilde{y}_i$ for each sample $x_i$ where the model prediction is correct when $\Tilde{y}_i$ is equal to the ground truth label $y_i$. The accuracy of a sample at epoch $t$ can be expressed as

\begin{equation}
    \label{eq:acc}
    c_i^t = \mathbf{1}_{\Tilde{y}^t_i = y_i}.
\end{equation}

Here, $\mathbf{1}_{\Tilde{y}^t_i = y_i}$ represents a binary variable that reduces to one if the model prediction is correct and zero otherwise. We further define a sample as ``forgotten" if the accuracy decreases in two subsequent epochs:
\begin{equation}
e_{i}^{t} = int( c^{t}_{i} < c^{t-1}_{i} ) \in {1, 0}
\end{equation}

Similar to \cite{toneva2018empirical}, we call the binary event $e_i^t$ a \emph{forgetting event}. Within the context of active learning, we quantify informativeness by the amount of forgetting events that occur for a given sample each active learning round. In contrast to existing literature, our definition of sample importance is not directly dependent on the data representation of the model but relies on artifacts from representation shifts. We reason that this characteristic is potentially favorable for out-of distribution active learning.

\begin{figure*}[!th]
    \begin{center}
        \includegraphics[scale=0.45]{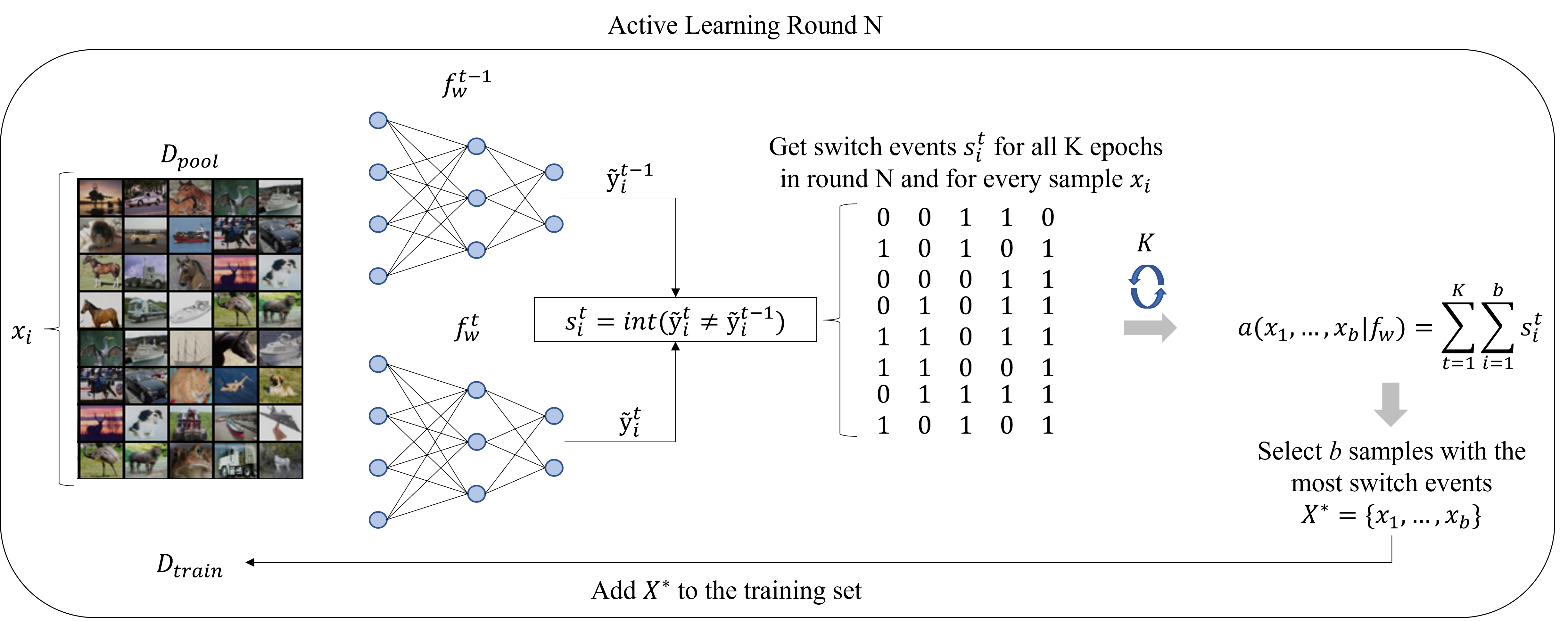}
    \end{center}
    \caption{Workflow of \texttt{FALSE}. Within the active learning round $N$, we gather switch events for every samples in $D_{pool}$ and aggregate them over all epochs $K$. Subsequently. we query $b$ samples with the most switch events.}
 \label{fig:flowchart}
\end{figure*}
\section{Forgetful Active Learning with Switch Events}
Even though forgetting events are simple to formulate and do not rely on the representation directly, the computation is not tractable in practice. Specifically, the accuracy evaluation in Equation~\ref{eq:acc} requires labels which are not present in $D_{pool}$ by definition. To alleviate this issue, we approximate forgetting events with prediction switches. A prediction switch occurs when the prediction of the model changes between two subsequent epochs. Formally, we write 

\begin{equation}
s_{i}^{t} = int( \Tilde{y}^{t}_i \neq \Tilde{y}^{t-1}_i ) \in {1, 0}.
\end{equation}

Similar to our previous definition, we call the binary event $s_i^t$ a \emph{switch event} at time $t$. Since samples with a larger amount of switch events are considered more informative, we select the batch $X^*$ with the most switch events in each round:

\begin{equation}
X^* = \argmax_{x_1, ..., x_b \in D_{pool}} \Sigma_{i=1}^b \Sigma_{t=1}^{K} s_i^t
\end{equation}

Here, $K$ refers to the total amount of epochs during an active learning round. We call our method \emph{Forgetful Active Learning with Switch Events} or \texttt{FALSE} in short. We show the workflow during each active learning round in Figure~\ref{fig:flowchart}. First, we count the switch events for every sample in $D_{pool}$ occurring within round $N$. Subsequently, we select $b$ samples with the most switch events and append them to the training set $D_{train}$.

%$\mathbf{Forgetful\ Active\ Learning\ with\ Switch\ Events}$ $\mathbf{FALSE}$

\begin{figure*}[!h]
    \begin{center}
    %\hspace*{-11mm}
        \includegraphics[scale=0.38]{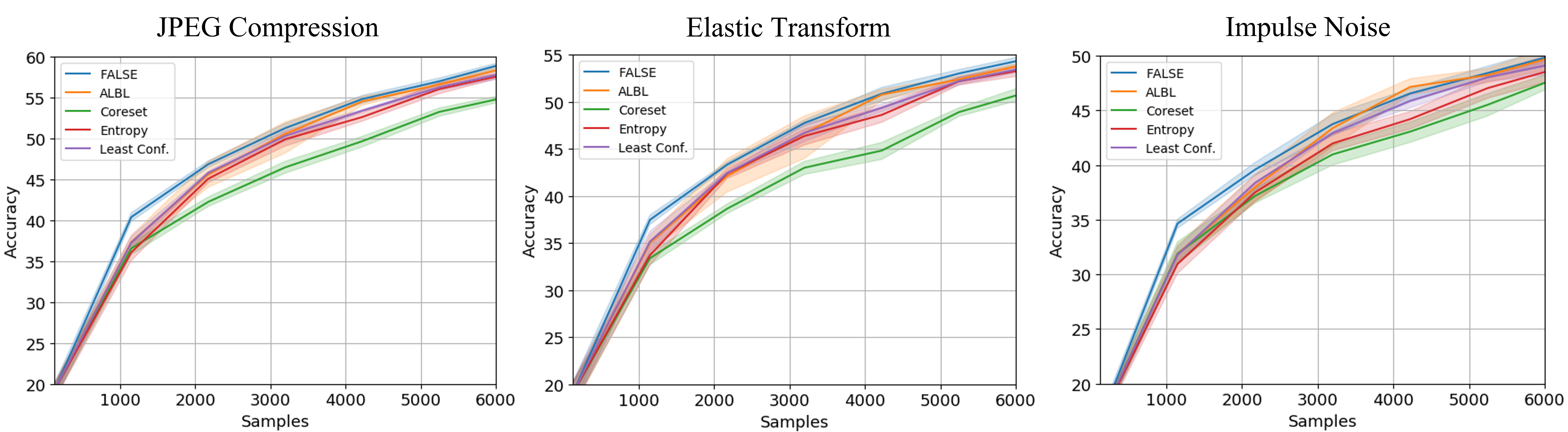}
    \end{center}
    \caption{\texttt{FALSE} in comparison to different query strategies on the level 2 CIFAR10-C \cite{hendrycks2019robustness} corruptions ``JPEG Compression", ``Elastic Transform", and ``Implulse Noise". We compare \texttt{FALSE} to active learning by learning (ALBL), coreset, least confidence sampling (Least Conf.), and entropy sampling (Entropy).}
 \label{fig:lcs-cifar10}
\end{figure*}
\begin{table}[!h]
	\centering
	\begin{tabular}{ |p{0.2cm}|p{1.9cm}||p{1.4cm}|p{1.4cm}|p{1.4cm}|}
		\hline
		
		\multicolumn{1}{|c|}{} & \multicolumn{1}{|c||}{Algorithms} & \multicolumn{1}{|c|}{CIFAR10-C}  & \multicolumn{1}{|c|}{CINIC10} & \multicolumn{1}{|c|}{CIFAR10}\\
		\hline
        \parbox[t]{2mm}{\multirow{5}{*}{\rotatebox[origin=c]{90}{ResNet-18}}}  	& \bf{FALSE}                           & $\mathbf{21.85}$ & $\mathbf{18.06}$ & $\mathbf{21.80}$ \\
                                                                        		& ALBL \cite{hsu2015active}       & 10.19            & 5.66             & 7.25             \\
                                                                        		& Coreset \cite{sener2017active}  & -41.87           & -35.07           & -41.04           \\
                                                                        		& Entropy \cite{wang2014new}      & 7.83             & 8.58             & 0.45             \\
                                                                        		& L. Conf. \cite{wang2014new}     & 16.23            & 13.47            & 9.93             \\
                                                                        		\hline
        \parbox[t]{2mm}{\multirow{5}{*}{\rotatebox[origin=c]{90}{DenseNet-121}}}  	& \bf{FALSE}                           & $\mathbf{13.92}$ & $\mathbf{16.50}$ & $\mathbf{15.47}$ \\
                                                                        		& ALBL \cite{hsu2015active}       & 7.69             & 6.67             & 3.28             \\
                                                                        		& Coreset \cite{sener2017active}  & -23.40           & -14.36           & -21.74           \\
                                                                        		& Entropy \cite{wang2014new}      & 1.77             & 2.52             & -7.76            \\
                                                                        		& L. Conf. \cite{wang2014new}     & -0.20            & 3.81             & -3.57            \\
                                                                        		\hline
        \parbox[t]{2mm}{\multirow{5}{*}{\rotatebox[origin=c]{90}{ResNet-34}}}  	& \bf{FALSE}                           & $\mathbf{22.20}$ & $\mathbf{23.30}$ & $\mathbf{27.45}$ \\
                                                                        		& ALBL \cite{hsu2015active}       & 7.56             & 9.67             & 13.96            \\
                                                                        		& Coreset \cite{sener2017active}  & -8.25            & -7.18            & -5.21            \\
                                                                        		& Entropy \cite{wang2014new}      & 14.80            & 15.31            & 12.51            \\
                                                                        		& L. Conf. \cite{wang2014new}     & 17.52            & 16.60            & 15.99            \\
		\hline
	\end{tabular}
	\caption{Area under difference curve with reference to random sampling over several datasets, query strategies, and architectures. Positive, and higher numbers are better. Negative numbers imply that the strategy has a lower accuracy curve than the random baseline.}
	\label{table:results-overall}
\end{table}
\section{Experiments}
\subsection{Experimental Setup}
We compare \texttt{FALSE} to four popular strategies in active learning literature: Entropy sampling \cite{wang2014new}, coreset \cite{sener2017active}, least confidence sampling \cite{wang2014new}, and active learning by learning \cite{hsu2015active}. We choose this constellation as it contains two strategies based on generalization difficulty (entropy and least confidence sampling), one based on data diversity (coreset), and a hybrid protocol that combines generalization difficulty and data diversity by switching between the coreset and least confidence strategy (active learning by learning). For our OOD experiments, we consider CIFAR10-C \cite{hendrycks2019robustness} (a corrupted version of CIFAR10), as well as CINIC10 \cite{cinic10} (a larger dataset with the same classes as CIFAR10). In addition, we validate on the CIFAR10 in-distribution test set. In all experiments, we train on the CIFAR10 training set. For CIFAR10-C, we test on all corruptions except "labels", "shot noise", and "speckle noise", and consider the difficulty levels 2 and 5. We choose this setup as it considers 17 realistic corruptions at an intermediate, as well as high difficulty. For CINIC10, we test on the CINIC10 test set. We start with an initial training pool size of randomly chosen 128 samples and query 1024 samples each round. Since query strategy performance can be sensitive to architecture choices, we perform our experiments on resnet-18, resnet-34, as well as densenet-121. We optimize with the adam variant of SGD with a learning rate of $1e-4$ and train our models until a training accuracy of 98\% is reached. Furthermore, we use pretrained model weights each round as this represents a realistic practical scenario where data may be redundant in one domain but scarce in another. Finally, we retrain our model form scratch each round to prevent warm starting \cite{ash2020warm}. Due to space limitations, we show the learning curves of three level 2 corruptions of the CIFAR10-C dataset in Figure~\ref{fig:lcs-cifar10}. Additionally, we summarize the performance of each protocol over the first 20 rounds by calculating the curve distance to randomly selecting samples each round (Table~\ref{table:results-overall}). For each experiment constellation, we subtract the accuracy curve of the respective strategy from the random selection baseline curve and calculate the area under the difference curve. In other words, the table shows the distance of each strategy to randomly selecting samples each round. A negative result implies that the strategy is overall worse than random selection, while a higher score means that the respective strategy outperforms random selection by a larger margin over the first 20 rounds. All results are averaged over five random seeds. In total, this amounts to 270 separate active learning experiments.

\subsection{Discussion}
From both Figure~\ref{fig:lcs-cifar10} and Table~\ref{table:results-overall} we observe that \texttt{FALSE} is a good choice for OOD active learning. In particular, we observe that \texttt{FALSE} is especially favorable in early rounds where few labeled data samples are available for training. We reason that the lack of training samples results in insufficiently calibrated representations and the importance rankings are more inaccurate when defined on the representation directly. For later rounds, enough training samples are available to learn a more conclusive representation and \texttt{FALSE} outperforms by a smaller margin (Figure~\ref{fig:lcs-cifar10}). On average, we observe up to 4.5\% improvements in accuracy across existing strategies. We further note, that \texttt{FALSE} is more robust to different datasets and architectures for both in-distribution and OOD experiments. In Table~\ref{table:results-overall}, we see that \texttt{FALSE} consistently outperforms random selection by a large margin over different datasets and architectures. This is evident in the consistent large area under the difference curve. In contrast several strategies, outperform random selection on some datasets or architectures but match or underperform the baseline in other settings. For instance, least confidence sampling exhibits a high area score on CIFAR10-C when resnet-34 is used, but slightly underperforms the random baseline with densenet-121. Furthermore, entropy sampling performs well on the in-distribution test set when resnet-34 is used but underperforms random by a significant margin when densenet-121 is used. We further observe, that \texttt{FALSE} also shows the strongest performance overall. This is evident in the largest area across all settings in Table~\ref{table:results-overall}.

\section{Conclusion}
In this paper, we introduced FALSE as a novel query strategy for OOD active learning. Specifically, we derived "informativeness" from the learning dynamics during training time. We approximated how often the network "forgot" unlabeled samples by counting switch events during each active learning round. We empirically analyzed our method with exhaustive experiments, and noted a clear improvement over existing methods in in-distribution, and out-of-distribution settings.

\vfill\pagebreak

\bibliographystyle{IEEEbib}
\bibliography{mybib}

\end{document}